\documentclass{article}
\usepackage{spconf,amsmath,graphicx}
\usepackage{multirow}
\usepackage{makecell}
\usepackage{amsfonts}
\usepackage{bm}
\usepackage[colorlinks,bookmarksopen,bookmarksnumbered,citecolor=green, linkcolor=red, urlcolor=blue]{hyperref}
\usepackage{bbding}

\title{Generalized Uncertainty-Based Evidential Fusion with Hybrid Multi-Head Attention for Weak-Supervised Temporal Action Localization}
\name{Yuanpeng He$^{1,\ast}$, Lijian Li$^{2,\ast}$, Tianxiang Zhan$^{2}$, Wenpin Jiao$^{1}$, Chi-Man Pun$^{2,\dagger}$ \thanks{$\ast$: Contributions of the authors are equal.}\thanks{$\dagger$: Corresponding author, E-mail: cmpun@umac.mo}}

\address{$^{1}$School of Computer Science, Peking University, Beijing, China;\\ Key Laboratory of High Confidence Software Technologies (Peking University), Ministry of Education
 \\
      $^{2}$ Department of Computer and Information Science, University of Macau, Macau, China}
\begin{document}
%
\maketitle
\begin{abstract}
Weakly supervised temporal action localization (WS-TAL) is a task of targeting at localizing complete action instances and categorizing them with video-level labels. Action-background ambiguity, primarily caused by background noise resulting from aggregation and intra-action variation, is a significant challenge for existing WS-TAL methods. In this paper, we introduce a hybrid multi-head attention (HMHA) module and generalized uncertainty-based evidential fusion (GUEF) module to address the problem. The proposed HMHA effectively enhances RGB and optical flow features by filtering redundant information and adjusting their feature distribution to better align with the WS-TAL task. Additionally, the proposed GUEF adaptively eliminates the interference of background noise by fusing snippet-level evidences to refine uncertainty measurement and select superior foreground feature information, which enables the model to concentrate on integral action instances to achieve better action localization and classification performance. Experimental results conducted on the THUMOS14 dataset demonstrate that our method outperforms state-of-the-art methods. Our code is available in \url{https://github.com/heyuanpengpku/GUEF/tree/main}.

\end{abstract}
\section{Introduction}
\label{sec:intro}
As one of the most essential tasks of video understanding, Temporal Action localization (TAL) targets at accurately positioning action boundaries, including start and end timestamps of action instances in untrimmed video and sorting out them. Numerous early studies \cite{DBLP:journals/ijcv/ZhaoXWWTL20, DBLP:conf/eccv/LinZSWY18, DBLP:conf/cvpr/LongYQTLM19} focus on utilizing fully-supervised methods to solve the task and achieve notable performance. Nevertheless, fully supervised methods necessitate a large number of videos with frame-wise annotations, which are time-consuming and labor-intensive. Furthermore, annotations provided by different annotators have distinct biases. Therefore, to address the aforementioned problems, various weakly supervised temporal action localization (WS-TAL) methods \cite{DBLP:conf/cvpr/Huang0022, DBLP:conf/eccv/ChenGYX22, DBLP:conf/wacv/ZhouW23, DBLP:conf/cvpr/LiCD0W023} have been proposed in recent years that only rely on lightly obtainable video-level labels, avoiding extensive annotation costs. 

Most weakly supervised methods adopt the localization-by-classification pattern that involves training an action classifier and applying it to generate a Class Activation Sequence (CAS), which consists of classification probabilities for snippets. Meanwhile, an attention branch is used to compute a weight sequence that represents the foreground probability of each snippet. Subsequently, a video-level prediction is obtained by aggregating the top-k snippets based on the attention sequence. However, for the sake of lacking fine-grained annotations, achieving great temporal action localization performance is still problematic with video-level labels. The primary challenge faced by current WS-TAL methods is action-background ambiguity. Specifically, during the training process, the classifier tends to pay attention to salient features, leading to the misclassification of background snippets as actions, while disregarding less prominent action snippets. This tendency enables the model to only take a fraction of action snippets rather than an entire action instance into consideration, resulting in inaccurate localization and classification results. Moreover, most existing WS-TAL methods directly employ a pre-trained I3D model to extract RGB and optical flow features, which encompass a large quantity of redundant information irrelevant to the task, consequently impeding performance.

To address the action-background ambiguity problem, we propose a Generalized Uncertainty-Based Evidential fusion (GUEF) module for WS-TAL, inspired by Traditional Evidential Deep Learning (TEDL), which is capable of obtaining the uncertainty of predictions by computing the uncertainty measure of each snippet-level evidence to filter useless background snippets. Within the GUEF module, the disturbance of background information is quantified by video-level uncertainty, which can adaptively eliminate redundant background snippets. Additionally, the snippet-level uncertainty can be spontaneously deduced to be video-level uncertainty. Therefore, the action-background ambiguity problem is alleviated effectively with this module. Besides, a Hybrid Multi-Head Attention (HMHA) is proposed, which effectively strengthens RGB and optical flow features extracted by the pre-trained I3D model by aligning their feature distribution with the WS-TAL task. 

In sum, the main contributions of this work are listed as follows:\par
(1) We propose a Generalized Uncertainty-Based Evidential Fusion module for the WS-TAL task, which can effectively eliminate action-background ambiguity problem by fusing snippet-level evidences.\par
(2) A Hybrid Multi-Head Attention module is proposed to enhance the extracted RGB and optical flow features, aligning the feature distribution more appropriately with the requirements of the WS-TAL task.\par
(3) The results of a large number of experiments conducted on the THUMOS14 dataset demonstrate the excellent performance of our proposed method, surpassing recent state-of-the-art methods.

\section{METHODOLOGY}

\subsection{Hybrid Multi-Head Attention}
\begin{figure}
    \centering
    \includegraphics[height=4.8cm, width=8.5cm]{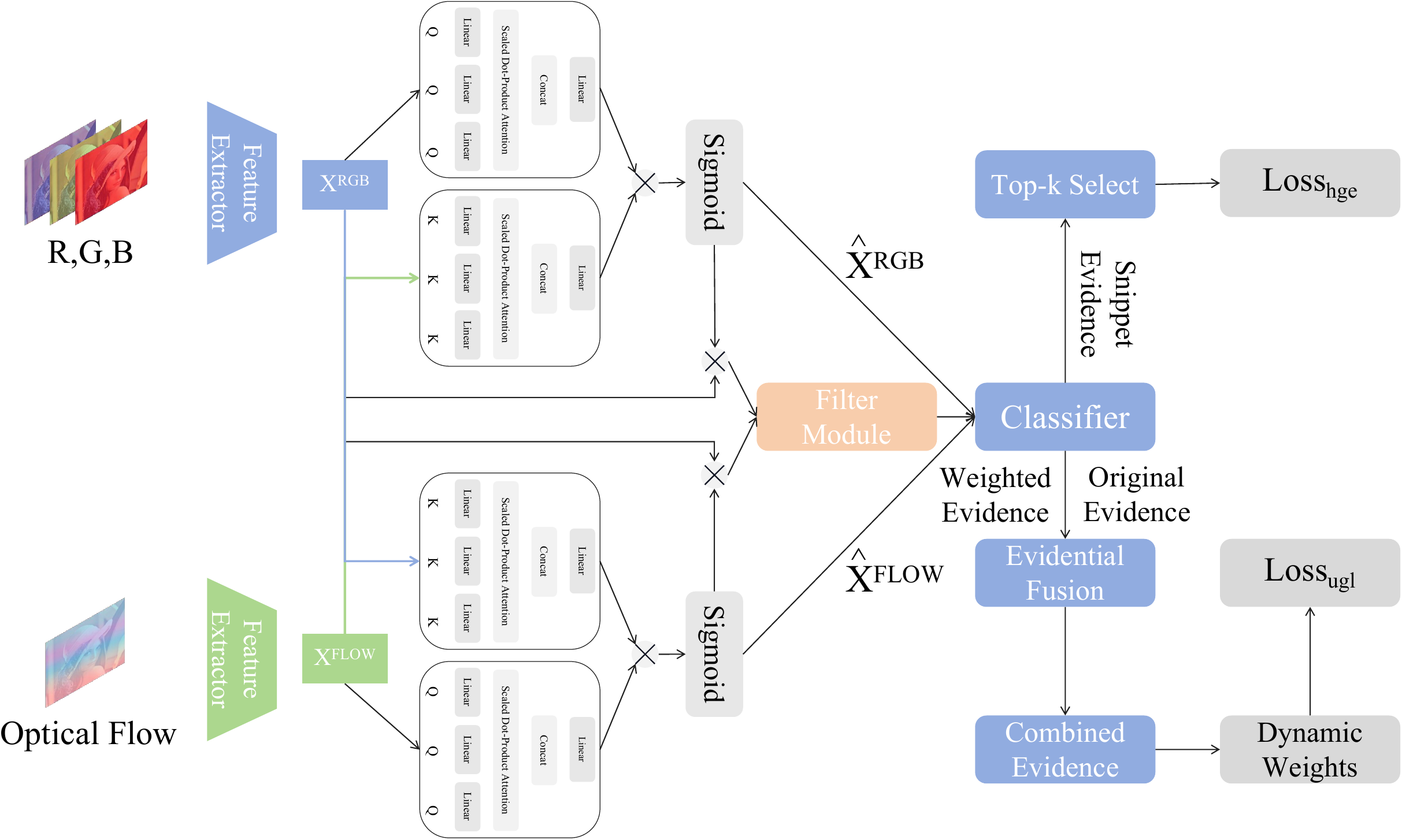}
    \caption{The overall structure of our model.}
    \label{model}
\end{figure}

A Hybrid Multi-Head Attention (HMHA), which consists of two sharing multi-head attention modules and a filtering module with an attention-like information interaction mechanism, is applied to enable the two modalities' weight distribution to approach each other. Following the existing methods, the untrimmed video $V$ is split into $W$ 16-frame snippets, which is non-overlapping, and a localization-by-classification strategy is adopted. The structure of this module is illustrated in Fig \ref{model}. Because both optical flow features and RGB features $X^{Flow}, X^{RGB} \in \mathbb{R}^{D \times W}$ are extracted by pre-trained models, i.e., I3D \cite{DBLP:conf/cvpr/CarreiraZ17}, there is numerous redundant information that is irrelevant to WS-TAL task in both features. Therefore, both of them are fed into two sharing multi-head attention modules $MHA$ to obtain two weights $A^{Flow}$, $A^{RGB} \in \mathbb{R}^{W}$, which are employed to eliminate task-irrelevant information contained in two initial features \cite{DBLP:conf/mm/HongFXSZ21}. The process can be stipulated as below:
 \begin{equation}
        A^{Flow}, A^{RGB}= MHA(X^{Flow}), MHA(X^{RGB}) 
\end{equation}
\begin{equation}
    \begin{split}
    \widehat{X}^{Flow} = {X}^{Flow} \otimes \sigma(A^{Flow} \otimes A^{RGB}) \\
    \widehat{X}^{RGB} = {X}^{RGB} \otimes \sigma(A^{RGB} \otimes A^{Flow})
    \end{split}
\end{equation}
where $A^{Flow}$ and ${A}^{RGB}$ can be regarded as ``query'' and ``key'' in multi-head attention module and $\sigma(\cdot)$ represents Sigmoid function. With optimized optical flow $\widehat{X}^{Flow}$ and RGB features $\widehat{X}^{RGB}$, we intend to use a filtering module $f_{attn}$, which consists of three 1D convolution layers and a sigmoid function, to extract their temporal attention weights. The process can be designed as below:
 \begin{equation}
        \widehat{A}^{Flow}, \widehat{A}^{RGB}= f_{attn}(\widehat{X}^{Flow}), f_{attn}(\widehat{X}^{RGB}) 
\end{equation}
After obtaining two optimized features and attention weights, we concatenate $\widehat{X}^{Flow}$ and $\widehat{X}^{RGB}$ as final output features. The final attention is an average of two generated attention weights. The process can be formulated as below:
\begin{equation}
    \begin{split}
        F = Concat(\widehat{X}^{Flow}, \widehat{X}^{RGB}) \\
        A = (\widehat{A}^{Flow} + \widehat{A}^{RGB}) / 2
    \end{split}
\end{equation}
Then, a class activation map (CAM) will be generated by a classifier containing three 1D convolution layers, whose input is the final optimized features $F$. And $A$ also can better guide the model to focus on foreground snippets because it fuses two modality attention weights.

\subsection{Deep Learning with Generalized Uncertainty-Based Evidential Fusion (GUEF)}
Evidential deep learning is developed on the basis of Subjective Logic-based Dempster-Shafer evidence theory \cite{DBLP:books/sp/Josang16, DBLP:series/sfsc/2008-219} which aims to avoid overconfidence of softmax-based classifiers on false predictions. Traditional evidential deep learning is designed under a frame of discernment which has $N$ mutually exclusive singletons $p_k, k = 1,...,N$ with corresponding belief mass $m(\{p_k\})$ and takes an overall uncertainty mass $U$ into consideration. Specifically, the sum of $N+1$ value of mass equals $1$, and each of them is non-negative, which can be defined as:
\begin{equation}
    \sum_{k = 1}^{N}m(\{p_k\}) + U = 1
\end{equation}
where $m_k \geq 0$ and $U \geq 0$. Assume $e_k \geq 0$ is an evidence corresponding to $k^{th}$ singleton, then mass of belief $m_k$ and uncertainty $U$ can be defined as:
\begin{equation}
    m(\{p_k\}) = \frac{e_k}{S}, U = \frac{N}{S}, S = \sum_{k=1}^{N}(e_k+1)
\end{equation}
However, according to the original definition of Dempster-Shafer evidence theory, the process of calculation of uncertainty in evidential deep learning is not precise enough \cite{DBLP:series/sfsc/Dempster08a} to fully represent the level of uncertainty of each complete piece of evidence. Here, we introduce the concept of multiple subsets into the traditional evidential deep learning, which can be defined as:
\begin{equation}
    m(\{\Theta\}) = U, \Theta = \{p_1, p_2, ..., p_N\} = p_{N+1}
\end{equation}
where the belief values of a multiplet $\Theta$ replace the position of the original uncertainty measure and unify uncertainty measures and belief values into one piece of newly-defined evidence \cite{DBLP:journals/ijon/TongXD21}. Then, for two pieces of evidences, the combination of them can be defined as:
\begin{equation}
\begin{split}
    m_{final}(\{p_k\}) = \frac{1}{1-Con}(m_1(\{p_k\}) * m_2(\{p_k\})\\ + m_1(\{p_k\}) * m_2(\{\Theta\}) + m_1(\{\Theta\}) * m_2(\{p_k\}))
\end{split}
\end{equation}
where $Con = \sum_{p_a \cap p_c = \emptyset} m_1(p_a)*m_2(p_c)$, which is called the coefficient of conflict between two pieces of evidences. To simplify the combination of evidences, the process is denoted as:
\begin{equation}
    e_{final} = e_1 \bigotimes e_2
\end{equation}
where $e_1$ and $e_2$ represent two complete pieces of evidences that contain all of the belief values of singletons and multiplets.

For the WS-TAL task, we are supposed to give a prediction about the video-level evidence $e_v = [e_{v,1}, ..., e_{v,T}\in \mathbb{R}]^{T}$ by using snippet-level evidence $e_s^1 = [e_{s,1}, ..., e_{s,T}]\in \mathbb{R}^{W \times T}$. After input of features flows through the proposed hybrid multi-head attention, an attention weight $A$, which is utilized to reconstruct the snippet evidence, can be acquired. Therefore, it is straightforward for us to obtain a new piece of evidence $e_s^2 = [e_{s,1}^2, ..., e_{s,T}^2]\in \mathbb{R}^{W \times T} = e_s^1 \cdot A$ with respect to snippet which is enhanced by the attention weights $A$. Based on the method of GUEF, we are able to fuse the original evidence and the reconstructed one. The process of combination can be defined as:
\begin{equation}
\begin{split}
    [e_s, m(\{\Theta\}] = (Concat[e_s^1/S,m_1(\{\Theta\}] \bigotimes \\Concat[e_s^2/S,m_2(\{\Theta\}]) \in \mathbb{R}^{W \times T}
\end{split}
\end{equation}
\begin{table*}[h]\small 
    \centering
    \renewcommand\arraystretch{1}
    \begin{tabular}{c|c|ccccccc|ccc}
        \hline
        \multirow{2}{*}{Supervision} & \multirow{2}{*}{Method} &  \multicolumn{7}{c|}{mAP@t-IoU(\%) $\uparrow$} & \multirow{2}{*}{\makecell{AVG \\ (0.1-0.5)}} & \multirow{2}{*}{\makecell{AVG \\ (0.3-0.7)}} & \multirow{2}{*}{\makecell{AVG \\ (0.1-0.7)}} \\
        \cline{3-9}
        & & 0.1 & 0.2 & 0.3 & 0.4 & 0.5 & 0.6 & 0.7 &  &  &  \\
        \hline
        \multirow{3}{*}{\makecell{Fully \\ (-)} } & SSN\cite{DBLP:journals/ijcv/ZhaoXWWTL20}(ICCV’17) & 60.3 & 60.3 &  50.6 & 40.8 & 29.1 & - & - & 49.6 & - & -  \\
        
        & BSN\cite{DBLP:conf/eccv/LinZSWY18}(ECCV’18) & - & - & 53.5 & 45.0 & 36.9 & 28.4 &  \textbf{20.0} & - & 36.8 & -  \\
        
        & GTAN\cite{DBLP:conf/cvpr/LongYQTLM19}(CVPR’19) & 69.1 & 63.7 &  57.8 & 47.2 & 38.8 & - & - & 55.3 & - & -  \\
        \hline
        \multirow{10}{*}{\makecell{Weakly \\ (I3D)} } 
        & FTCL \cite{DBLP:conf/cvpr/GaoCX22}(CVPR’22) & 69.6 & 63.4 & 55.2 & 45.2 & 35.6 & 24.3 & 12.2 & 53.8 & 34.4 & 43.6  \\
        
        & Huang et al.\cite{DBLP:conf/cvpr/Huang0022}(CVPR’22) & 71.3 & 65.3 & 55.8 & 47.5 & 38.2 & 25.4 & 12.5 & 55.6 & 35.9 & 45.1  \\
        
        & ASM-Loc\cite{DBLP:conf/cvpr/HeYKCZS22}(CVPR’22) & 71.2 & 65.5 & 57.1 & 46.8 & 36.6 &  25.2 & 13.4 & 55.4 & 35.8 & 45.1 \\
        
        
        & Li et al.\cite{DBLP:conf/mm/LiGYC22}(MM’22) & 69.7 & 64.5 & 58.1 & 49.9 & 39.6 &  27.3 & 14.2 & 56.3 & 37.8 & 46.1  \\
        
        & DELU\cite{DBLP:conf/eccv/ChenGYX22}(ECCV’22)  & 71.5 & 66.2 & 56.5 & 47.7 & 40.5 &  27.2 & 15.3 & 56.5 & 37.4 & 46.4 \\
        
        & TFE-DCN\cite{DBLP:conf/wacv/ZhouW23}(WACV’23)  & 72.3 & 66.5 & 58.6 & 49.5 & 40.7 &  27.1 & 13.7 & 57.5 & 38.0 & 46.9 \\
        
        & Wang et al.\cite{DBLP:conf/cvpr/WangLW23}(CVPR’23) & 73.0 & 68.2 & 60.0 & 47.9 & 37.1 & 24.4 & 12.7 & 57.2 & 36.4 & 46.2 \\

        & Li et al.\cite{DBLP:conf/cvpr/LiCD0W023}(CVPR’23) & - & - & 56.2 & 47.8 & 39.3 & 27.5 & 15.2 & - & 37.2 & - \\
        
        & Ren et al.\cite{DBLP:conf/cvpr/RenYZ023}(CVPR’23) & 71.8 & 67.5 & 58.9 & 49.0 & 40.0 & 27.1 & 15.1 & 57.4 & 38.0 & 47.0 \\
       

        
        \cline{2-12}
        & Ours & \textbf{74.5} & \textbf{69.1} & \textbf{60.3} &\textbf{51.2} &\textbf{42.1} & \textbf{29.4} & 15.5 &\textbf{59.5} &  \textbf{39.7}& \textbf{48.9}\\   
        \hline
    \end{tabular}
    \caption{The experimental results of our model on THUMOS 14 compared with state-of-the-art methods}
    \label{THUMOS}%
\end{table*}
\subsection{Learning and Inference}
A classifier $\mathbf{C}$ is utilized to predict class activation sequence (CAS) over snippet-wise features $F = [\mathbf{f_1},...,\mathbf{f_W}] \in \mathbb{R}^{D \times W}$ and $D$ is dimension of features, which is given as $z = [z_1,...,z_W] \in \mathbb{R}^{W \times (T+1)}$. Moreover, an attention score $A=[A_i,...,A_W] \in \mathbb{R^{W}}$ is produced by the hybrid multi-head attention, which represents the possibility that snippets belong to the foreground. Then, the overall classification probability $\hat{p}$ can be obtained by aggregating CAS, which utilizes the most preferable L snippets according to attention score $A$. The process of aggregation can be defined as:
\begin{equation}
    \hat{p} = \frac{1}{L}\sum_{t \in \xi, |\xi| = L}^{\xi= \arg\max_{\xi}\sum_{t\in \xi}A_t}z_{t} \label{eq:9}
\end{equation}
where $z_t = C(F)$. On the basis of prediction $\hat{p}$, we optimize it by comparing it with ground-truth label $p$:
\begin{equation}
    \mathcal{L}_{cla} = Cross\_entropy(\hat{p}, p)
\end{equation}

Then, considering the attention score $A$, which indicates the probability of snippets belonging to the foreground and probability $z_{t,T+1}$ in $z\in \mathbb{R}^{W \times (T+1)}$ represents background probability of $t^{th}$ snippet. It is intuitive to consider that $A$ and $z_{t,T+1}$ are complementary which are controlled by uncertainty measures produced by generalized evidential fusion:
\begin{equation}\footnotesize
    \mathcal{L}_{\mu ef} = (\Delta\cdot tanh (\sigma(h)\varphi(m_s(\{\Theta)\}) + 1)\sum_{t=1}^{W} L1_{norm}|1-A_t-z_{t,T+1}|
\end{equation}
where $m_s(\{\Theta)\}$ is acquired by sorting snippet-level uncertainty $m(\{\Theta)\}$ from each evidence in a descending order and $\sigma(h) = \frac{2h}{H} - 1 \in[-1,1], h=1,...,H$. More specifically, $h$ represents the current epoch reference, $H$ denotes the total number of training epochs. Moreover, $\varphi(m_s(\{\Theta)\} = \frac{2m_s(\{\Theta)\}}{W} -1 \in [-1,1], m_s = 1,...,W$ and $\Delta$ controls amplitudes of changes of $tanh (\sigma(h)\varphi(m_s(\{\Theta)\}$. Based on the traditional evidential deep learning (TEDL), with respect to a certain sample $X$, the corresponding Dirichlet distribution can be obtained as:
\begin{equation}
    D(q|\alpha) = \left\{
\begin{aligned}
\frac{1}{B(\alpha)}\prod_{j=1}^{T}q_j^{\alpha_j -1}, for\ q \in \mathbb{S}_T \\
0, \ Otherwise \qquad \\
\end{aligned}
\right.
\end{equation}
where $\alpha_j$ corresponds to $e_j+1, j = 1,...,T$, $T$ is the number of classes. Besides, $q$ is a point on the T-dimensional unit simplex $\mathbb{S}_T$ \cite{DBLP:conf/nips/SensoyKK18}. With respect to snippet-level evidence, we refined the label vector by utilizing belief values from evidence and assigning smaller weights to samples that possess higher levels of uncertainty, which is designed to lead the model to pay less attention to background noise and focus on truly important and practical features. Following the classical optimization algorithm of TEDL \cite{DBLP:conf/nips/SensoyKK18} and utilizing newly proposed hybrid attention, a novel method to adapt to the tasks of WS-TAL is defined as:
\begin{equation}\small
    \mathcal{L}_{hge} = \sum_{i=1}^{M}(1 - U_{e_s^1}) \sum_{j=1}^{T}\frac{p_j^{(i)}/e_j^{1,(i)}}{\sum_{j=1}^{T}p_j^{(i)}/e_j^{1,(i)}}(logS^{(i)}-log\alpha_j^{(i)})
\end{equation}
where $U_{e_s^1}$ is obtained from original snippet-level evidence, video-level evidence can be acquired by aggregating snippet-level evidences using eq.\ref{eq:9}. Synthesizing the designed targets, we are able to acquire the final loss function which can be defined as:
\begin{equation}
    \mathcal{L} = \mathcal{L}_{cla} + \lambda_1 \mathcal{L}_{\mu gl} + \lambda_2 \mathcal{L}_{hge}
\end{equation}
where $\lambda_1$ and $\lambda_2$ are trade-off parameters.
\section{Experiments and Results}

\subsection{Datasets and Metrics}
We conduct a large amount of experiments to evaluate the proposed method on THUMOS14 \cite{THUMOS14} dataset. THUMOS14 is composed of 200 validation videos and 213 testing videos with 20 action classes. Besides, The mean Average Precision (mAP) with different Intersection-over-Union (IoU) thresholds, which is regarded as a standard evaluation metric for WS-TAL tasks, is used to evaluate the performance of the proposed model.
\begin{table}[h]\footnotesize
\setlength{\tabcolsep}{1.8mm}{
    \centering
    \begin{tabular}{c|c|c|ccccc}
        \hline
        \multirow{2}{*}{Exp}  & \multirow{2}{*}{$GUEF$} & \multirow{2}{*}{$HMHA$} & \multicolumn{5}{c}{mAP@IoU(\%) $\uparrow$} \\
        \cline{4-8}
        & & & 0.1 & 0.3 & 0.5 & 0.7 & \makecell{AVG \\ (0.1-0.7)} \\
        \hline
        1 & \XSolidBold & \XSolidBold & 72.1 & 56.7 & 38.7 & 12.3 & 46.5 \\
        2 & \XSolidBold & \CheckmarkBold & 72.9 & 58.3 & 40.2 & 13.7 & 47.3 \\
        3  & \CheckmarkBold & \XSolidBold & 73.9 & 60.0 & 41.8 & 14.5 & 48.4 \\
        \hline
        4 & \CheckmarkBold & \CheckmarkBold & \textbf{74.5} & \textbf{60.3} & \textbf{42.1} & \textbf{15.5} & \textbf{48.9} \\
        \hline
    \end{tabular}}
    \caption{The experimental results of ablation study}
    \label{table2}
\end{table}
\subsection{Experiments Setup}
Following existing WS-TAL methods, we employ the pre-trained I3D model to extract optical flow features and RGB features \cite{DBLP:conf/cvpr/CarreiraZ17}. In addition, the batch size is set to 10. An Adam optimizer with a learning rate of $5e-5$ is employed to optimize our model on the THUMOS14 dataset for 5000 iterations. For hyperparameters $\lambda_1$ and $\lambda_2$, we set it to 0.8, 1. During training, the maximum number of sample snippets on THUMOS14 is set to 320 and the amplitude $\Delta$ is set to 0.7.

\subsection{Performance Comparison with State-of-the-Art Models}
Table \ref{THUMOS} indicates the performance comparison between fully and weakly supervised methods and our proposed model on the THUMOS14 dataset. Our method achieves highest performance on mAP with IoU (0.1-0.6) and mAP@AVG. Only a fully-supervised method BSN achieves best performance 20\% on mAP@0.7. All of the results demonstrate the state-of-the-art performance of our method for the WS-TAL task.
\subsection{Ablation Study}
As is shown in Table \ref{table2}, $HMHA$ denotes the hybrid multi-head attention module. $GUEF$ denotes the proposed generalized uncertainty-based evidential fusion. Here, we explore the effectiveness of two mentioned modules on the THUMOS14 dataset. Notably, the omission of $GUEF$ results in significant performance degradation, confirming its effectiveness. Besides, only removing $HMHA$, the performance also shows a moderate degradation within 0.6. Therefore, Table \ref{table2} evidently indicates that each component of our model makes a tremendous contribution to improving performance.

\section{Conclusion}
In this paper, we propose a generalized uncertainty-based evidential fusion and hybrid multi-head attention module, which effectively eliminates action-background ambiguity and filters redundant information from pre-trained features to enable the model to focus on foreground snippets, consequently improving performance. Experimental results on the THUMOS14 dataset compared with the latest state-of-the-art methods demonstrate the effectiveness of our proposed method. Considering the fact that pseudo-label is also efficacious on WS-TAL tasks, we will conduct further research.
\bibliographystyle{IEEEbib}
\bibliography{strings,refs}

\end{document}